\documentclass[runningheads]{llncs}
\usepackage[T1]{fontenc}
\usepackage{graphicx,verbatim}
\usepackage{amsfonts}
\usepackage[table,xcdraw]{xcolor}
\usepackage{colortbl}
\usepackage{float}
\usepackage{graphicx}
\usepackage{siunitx}
\usepackage{pifont}
\usepackage{authblk}
\usepackage{url}
\usepackage{textcomp}
\usepackage[T1]{fontenc}
\usepackage{xcolor}
\usepackage{graphicx,verbatim}
\usepackage{mathtools, amssymb} 
\usepackage{booktabs}
\usepackage[misc]{ifsym}
\usepackage{makecell}
\usepackage{courier}     
\usepackage{bm}
\usepackage{multirow}
\usepackage{colortbl}
\usepackage[table]{xcolor}
\usepackage{cite}
\usepackage{hyperref}
\title{UniSpine-GS: An Efficient Physics-Aware Gaussian Framework for Cross-Modality Multi-view Spine Image Synthesis}
\author{Qiuhua Chen\inst{1}\and
        Changning Yu\inst{1}
       \and
        Na Huang\inst{2}\and
        Chao Sun\inst{1,3}$^{(\textrm{\Letter})}$\and
        Bo Du\inst{1,3,4,5}\\
}
\authorrunning{Q. Chen, C. Yu et al.} 

\institute{
School of Computer Science, Wuhan University, Wuhan, China 
\\
\and
The Department of Ultrasound of Renmin Hospital East Branch of Wuhan University, Wuhan, China
\\
\and
Institute of Artificial Intelligence, Wuhan University, Wuhan, China
\\
\and
National Engineering Research Center for Multimedia Software, Wuhan University, Wuhan, China
\\
\and
Hubei Key Laboratory of Multimedia and Network Communication Engineering, Wuhan University, Wuhan, China
\\
\email{chaosun@whu.edu.cn}}
\begin{document}
\maketitle
\markright{UniSpine-GS}
\begin{abstract}

The diagnosis of spinal diseases is often assisted by 3D imaging techniques in clinical practice. However, precise 3D spinal assessment is limited by the high costs of 3D imaging hardware and the challenges posed by the physical differences between imaging modalities, which hinder the generalizability of models. To address these issues, we propose \textbf{UniSpine-GS}, an efficient, physics-aware Gaussian framework designed for novel-view projection rendering in multi-view spine imaging via a 3D-aware representation. Instead of performing explicit 3D reconstruction, our approach learns a geometry-aware Gaussian representation that ensures anatomical consistency across different views. We introduce SPWM, a structure-guided loss reweighting strategy to improve boundary fidelity and local details. We evaluate our method on the CTSpine3D dataset and a newly constructed 3D fetal ultrasound dataset, FeSpine3D. Our results demonstrate that UniSpine-GS significantly outperforms existing methods across all metrics, offering a practical and cost-effective solution for unified multi-view medical imaging. Our code is publicly available at \href{https://github.com/orangeisland66/UniSpine-GS}{https://github.com/orangeisland66/UniSpine-GS}.

\keywords{Multi-view Spine Imaging \and Cross-Modality 3D Synthesis \and X-Gaussian Representation \and Medical Image Synthesis}

\end{abstract}
\section{Introduction}
Spinal disorders such as scoliosis, vertebral degeneration, and spondylolisthesis require an accurate assessment of spinal anatomy in three dimensions (3D) for reliable diagnosis and treatment planning~\cite{kumar2024review}. In clinical practice, X-ray imaging is widely used due to its low cost, low dose, and broad accessibility, while ultrasound is highly valuable for bedside examinations and obstetric scenarios because it is radiation-free, portable, and real-time. However, accurate 3D spinal assessment remains challenging. High-quality 3D imaging and dedicated acquisition systems are expensive and not always available. Moreover, large differences in imaging mechanisms and visual appearance across modalities make it difficult for models to maintain structural consistency and generalization in multi-modality and multi-view settings~\cite{zhou2023uxdiff,kazerouni2023diffusion}. These limitations motivate the need for cost-effective methods that enable cross-view and cross-modality spine image synthesis with reliable structural representations.

Recent advances in neural fields and differentiable rendering have enabled learning 3D representations from limited views for novel-view synthesis. NeRF~\cite{mildenhall2021nerf} achieves high fidelity but is often computationally expensive, motivating accelerated neural-field variants based on efficient encodings and anti-aliasing designs~\cite{muller2022instant,barron2023zip}. In contrast, 3D Gaussian Splatting (3DGS) uses explicit Gaussian primitives with highly parallel rasterization to substantially improve training and rendering efficiency~\cite{kerbl20233d,bao20253d}. For projection imaging, radiative Gaussian methods integrate radiative forward models with Gaussian representations~\cite{x_gaussian} and have been extended to rectified tomographic reconstruction and continuous-time 4D (dynamic) tomographic reconstruction~\cite{r2_gaussian,x2_gaussian}, while structure-aware priors improve stability under sparse-view conditions~\cite{sax_nerf}. Beyond static synthesis, Gaussian-based representations have been explored for SLAM, dynamic scenes, and inverse rendering~\cite{yugay2023gaussian,wu20244d,Liang_2024_CVPR}, and recent studies investigate SfM-free initialization and compact parameterizations~\cite{fu2024colmap,deng2024compact} as well as generative Gaussian modeling from limited observations~\cite{tang2023dreamgaussian}. Meanwhile, attention-enhanced or modality-enhanced neural fields improve robustness under non-ideal settings~\cite{feng2025ae,liu2025dsem}. Nevertheless, cross-modality multi-view spine synthesis remains underexplored, particularly for ultrasound, whose reflection/scattering-driven formation differs fundamentally from X-ray transmission, making faithful acoustic simulation costly.

Building on these ideas, we propose UniSpine-GS, an efficient and physics-aware Gaussian framework for cross-modality, multi-view 2D spine image synthesis. UniSpine-GS optimizes a geometry-aware Gaussian representation and enforces cross-view consistency through a unified forward operator. For ultrasound, we use a pragmatic proxy by treating the 3D volume as a projectable scalar field and generating multi-view images using the same differentiable digitally reconstructed radiograph (DRR) operator, without claiming a physically exact acoustic model. To improve fine-grained anatomy under sparse views, we introduce a Structure Prior Weight Map (SPWM) that reweights the loss to emphasize vertebral boundaries and texture-informative regions, with warm-up scheduling for stable optimization.

Our contributions are summarized as follows:
\begin{itemize}
\item We propose UniSpine-GS, an efficient physics-aware Gaussian framework for cross-modality multi-view 2D spine image synthesis.
\item We introduce an SPWM, a parameter-light structure-guided loss reweighting strategy with warm-up scheduling to enhance boundary fidelity and local details under sparse-view supervision.
\item We construct the FeSpine3D dataset and validate UniSpine-GS on CTSpine3D and FeSpine3D, achieving consistent improvements in structural consistency and local detail with high efficiency.
\end{itemize}

\section{Methodology}
The overall architecture of UniSpine-GS is a differentiable radiative rendering pipeline with an explicit 3D Gaussian representation. 
Given multi-view observations and known acquisition parameters, we initialize and optimize a set of Gaussians and render a predicted projection for a sampled viewpoint via a unified radiative forward operator. 
Training minimizes a hybrid objective combining an SSIM loss and a weighted Charbonnier loss reweighted by SPWM. 
Gaussian parameters are updated through backpropagation with a densification–pruning schedule to progressively refine the representation.
\begin{figure}
    \centering
    \includegraphics[width=1\linewidth]{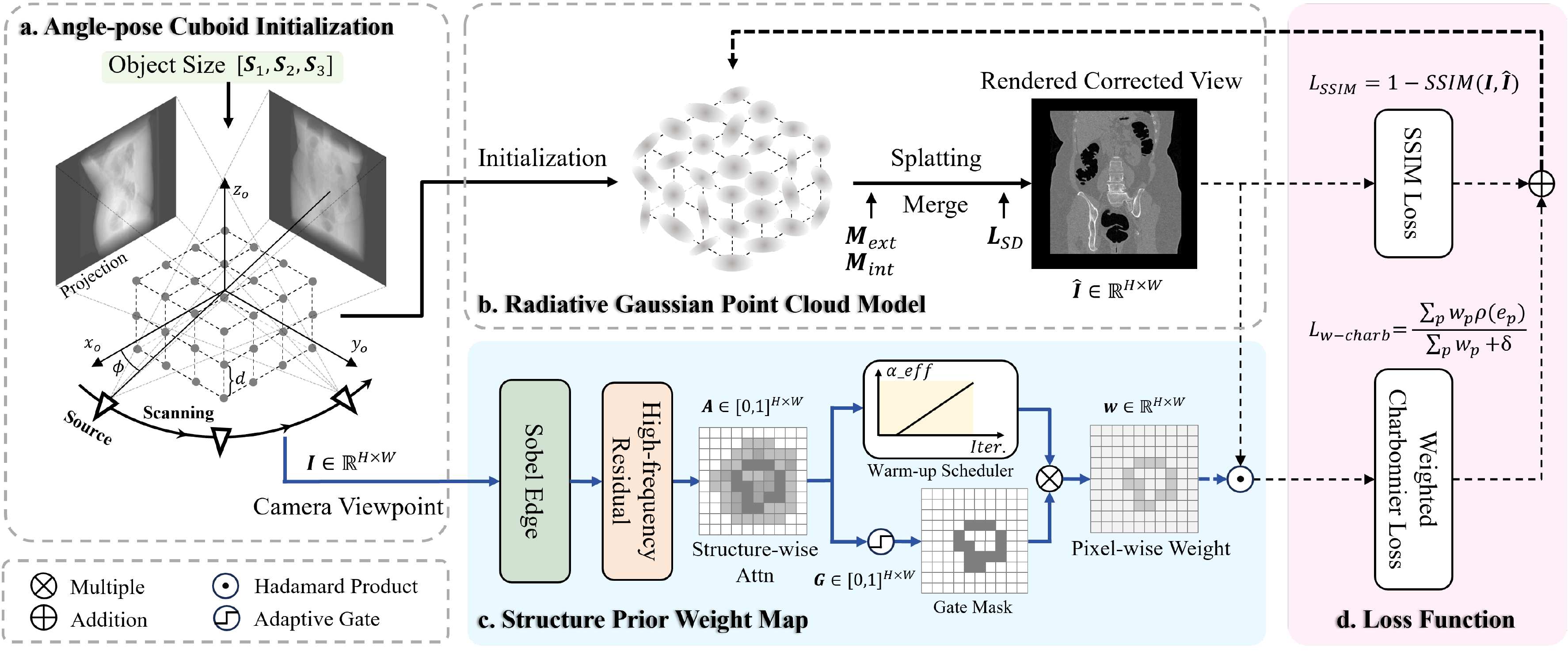}
    \caption{Overview of UniSpine-GS. We optimize a radiative Gaussian representation for multi-view spine synthesis using a unified forward operator. An SPWM derived from the Camera Viewpoint image reweights the reconstruction loss, together with an SSIM term, to emphasize structurally informative regions.}
    \label{fig:enter-label}
\end{figure}
\subsection{Angle-pose Cuboid Initialization (ACUI)}

Efficient optimization requires reliable initialization of camera poses and Gaussian centers. Unlike natural-scene 3DGS pipelines that depend on SfM, projection imaging typically provides scanner geometry, yet low contrast and transmission superposition make feature matching unreliable.
We employ the ACUI strategy introduced in X-Gaussian~\cite{x_gaussian} for circular cone-beam acquisition. Camera intrinsics and extrinsics are computed analytically from scanner parameters, and the projection matrix is determined for each azimuth angle $\phi$ from the known geometry. Gaussian centers are initialized by uniformly sampling 3D points within a cuboid that encloses the target anatomy at an interval $d$. The cuboid is centered at the world origin and specified by dimensions $\textbf{\textit{S}}_1$, $\textbf{\textit{S}}_2$, and $\textbf{\textit{S}}_3$, while covariance, opacity, and radiative features are randomly initialized. This geometry-driven initialization avoids costly preprocessing and provides a structured spatial prior, improving stability under sparse-view supervision.
\subsection{Radiative Gaussian Point Cloud Model}

Inspired by X-Gaussian~\cite{x_gaussian}, our foundational model formulates radiative Gaussian splatting for projection imaging.
An object is represented as a set of 3D Gaussian primitives: 
\[
\mathcal{G}=\{(\boldsymbol{\mu}_i,\mathbf{\Sigma}_i,\alpha_i,\mathbf{f}_i)\}_{i=1}^{N_p},
\tag{1}
\]
where $\boldsymbol{\mu}_i$ denotes the Gaussian center, $\mathbf{\Sigma}_i$ the covariance, $\alpha_i$ the opacity, and $\mathbf{f}_i$ the learnable radiative feature. 

Following X-Gaussian, we model a view-independent radiation intensity using a Radiation Intensity Response Function (RIRF):
\[
\mathbf{i}_i=\mathrm{Sigmoid}(\boldsymbol{\lambda}\odot \mathbf{f}_i),
\tag{2}
\]
where $\boldsymbol{\lambda}$ is a fixed weight vector and $\odot$ denotes element-wise multiplication.

Given camera extrinsics $\mathbf{M}_{ext}$ and intrinsics $\mathbf{M}_{int}$, we render the projection via a differentiable radiative forward operator based on transmittance-driven compositing along each ray:

\begin{gather}
\hat{\bm{\mathit{I}}} = F_{\mathrm{DRR}}(\mathbf{M}_{ext},\mathbf{M}_{int},\mathcal{G}), \tag{3}\\
\hat{\bm{\mathit{I}}}(p) = \sum_{j} T_j(p)\,\sigma_j(p)\,\mathbf{i}_j, \tag{4}\\
T_j(p) = \prod_{k<j}\bigl(1-\sigma_k(p)\bigr), \tag{5}
\end{gather}
where $\sigma_j(p)$ denotes the contribution of the $j$-th Gaussian at pixel $p$, and $T_j(p)$ is the accumulated transmittance. This formulation provides an efficient and physically consistent forward model~\cite{x_gaussian,r2_gaussian}.

Building on this unified radiative formulation, UniSpine-GS introduces a structure-aware strategy to enhance boundary fidelity and consistency.
\subsection{Structure Prior Weight Map (SPWM)}

To improve fine-grained anatomical reconstruction, we introduce an SPWM that emphasizes informative regions during training.

\subsubsection{Structure-wise Attention Map}

Given the ground-truth (GT) projection used only during training $\bm{\mathit{I}} \in \mathbb{R}^{H \times W}$, we compute a parameter-light Structure-wise Attention Map $\bm{\mathit{A}} \in [0,1]^{H \times W}$. 
We extract Sobel gradients and high-frequency cues via Gaussian-blur residuals. 
We fuse them into an attention map and normalize it to $[0,1]$:

\[
\bm{\mathit{A}} = \mathrm{Normalize}\big(E + \lambda_{\mathrm{hf}} H\big),
\tag{6}
\]
where $E$ denotes gradient magnitude, $H$ the high-frequency residual, and $\lambda_{\mathrm{hf}}$ balances their contributions. 
This Structure-wise Attention Map highlights vertebral boundaries and texture-rich regions.

\subsubsection{Pixel-wise Weight Map}

Based on the Structure-wise Attention Map, we compute a Pixel-wise Weight Map $\bm{\mathit{w}} \in \mathbb{R}^{H \times W}$ using a warm-up scheduler and an adaptive quantile threshold with a soft Gate Mask.
A linear warm-up factor $r(t)$ progressively increases the weighting strength after iteration $t_s$. 
An adaptive threshold $\tau$ is computed as the $q$-quantile of $\bm{\mathit{A}}$ to suppress noisy responses. 
The final weights are defined as:

\[
\bm{\mathit{w}} = 1 + \alpha_{\mathrm{eff}} \bm{\mathit{G}} \odot \bm{\mathit{A}},
\tag{7}
\]
where $\alpha_{\mathrm{eff}} = \alpha r(t)$ and $\bm{\mathit{G}}$ is the soft gate mask derived from $\tau$. 
This design emphasizes informative regions while maintaining stable optimization dynamics during early-stage training.

\subsection{Loss Function}

Given a sampled viewpoint, the unified renderer produces a Rendered Corrected View $\hat{\bm{\mathit{I}}}$, which is supervised by the ground-truth image $\bm{\mathit{I}}$. We optimize the Gaussian parameters using a hybrid loss that combines an SSIM term with a Weighted Charbonnier reconstruction term reweighted by the SPWM:
\begin{gather}
\mathcal{L}
=
(1-\lambda_{\mathrm{dssim}})\,\mathcal{L}_{w\text{-}\mathrm{charb}}
+
\lambda_{\mathrm{dssim}}\left(1-\mathrm{SSIM}\right),
\tag{8}\\
\mathcal{L}_{w\text{-}\mathrm{charb}}
=
\frac{\sum_{p} w(p)\,\rho(e(p))}
{\sum_{p} w(p)+\delta}.
\tag{9}
\end{gather}
Here, $\rho(\cdot)$ denotes the Charbonnier penalty, $e(p)$ is the per-pixel residual, and $w(p)$ is the SPWM weight at pixel $p$. This hybrid formulation improves boundary sharpness, suppresses low-contrast artifacts, and enhances cross-view structural consistency while maintaining stable optimization.

\section{Experimental Results}
\subsection{Dataset and System Implementation}
\subsubsection{Dataset}
We employ two 3D spine datasets for cross-modality multi-view synthesis. CTSpine3D~\cite{deng2021ctspine1k,yu2025veganet} contains over 600 CT volumes, and FeSpine3D contains 100 3D ultrasound volumes. For each volume, we render multi-view projections using a GPU-accelerated differentiable DRR operator (implemented with TIGRE~\cite{biguri2016tigre} for efficiency) under known cone-beam geometry. In particular, FeSpine3D is derived from clinically acquired fetal spine ultrasound data collected under routine clinical protocols, and all data were de-identified and anonymized prior to use.
\subsubsection{Implementation Details}
Our model is implemented in PyTorch, following the training and densification--pruning pipeline of X-Gaussian~\cite{x_gaussian}. All experiments are conducted on a single NVIDIA RTX 3090 GPU. We use Adam with $\beta_1=0.9$, $\beta_2=0.999$, and $\epsilon=10^{-15}$, and train for 20{,}000 iterations. The learning rates for the Gaussian attributes are set to 0.002 for the feature parameters, 0.008 for opacity, 0.005 for scaling, and 0.001 for rotation.

We enable the proposed structure-prior weighting with $\alpha=0.8$, warm-up starting at iteration 2{,}000 for 5{,}000 iterations, and construct the attention map using $\lambda_{\mathrm{hf}}=0.5$ with a $3\times3$ Gaussian blur kernel ($\sigma=1.5$); we adopt adaptive gating with quantile $q=0.85$. Densification is applied from iteration 500 to 8{,}000 every 200 iterations with a gradient threshold of $1.5\times10^{-4}$.
\subsubsection{Evaluation Metrics}We evaluate UniSpine-GS in terms of both reconstruction quality and computational efficiency. 
For image quality, we report Peak Signal-to-Noise Ratio (PSNR) and Structural Similarity Index Measure (SSIM) on held-out views, following common practice in novel-view synthesis and sparse-view reconstruction~\cite{mildenhall2021nerf,zha2022naf,chen2022tensorf,x_gaussian}. 
For efficiency, we measure inference speed in frames per second (fps) during rendering and report end-to-end training time (min/s) under the same hardware setting. All results are averaged over test views. Higher is better for PSNR/SSIM/fps, and lower is better for training time.
\subsection{Comparison Experiments}

\begin{table*}[t]
\centering
\caption{Quantitative comparison of reconstruction quality and efficiency on CTSpine3D and FeSpine3D datasets. Best results in \textbf{bold}, while second best \underline{underlined}.}
\setlength{\tabcolsep}{1.73mm}{
\begin{tabular}{llcccc}
\hline
\textbf{Dataset} & \textbf{Method} & \makecell{\textbf{Infer Speed}\\\textbf{(fps)$\uparrow$}} & \makecell{\textbf{Train Time}\\\textbf{(min)$\downarrow$}} & \textbf{PSNR$\uparrow$} & \textbf{SSIM$\uparrow$} \\
\hline
\multirow{5}{*}{\textbf{CTSpine3D}} 
& IntraTomo~\cite{zang2021intratomo} & 0.27 & \underline{33}  & 35.27 & 0.9641 \\
& NeRF~\cite{mildenhall2021nerf}     & 0.02 & 410 & \underline{37.61} & \underline{0.9802} \\
& TensoRF~\cite{chen2022tensorf}     & 0.06 & 185 & 35.60 & 0.9600 \\
& NAF~\cite{zha2022naf}              & \underline{0.37} & 93 & 36.64 & 0.9741 \\
\rowcolor{gray!10}
& Ours                               & \textbf{113.14} & \textbf{15} & \textbf{46.54} & \textbf{0.9938} \\
\hline
\multirow{5}{*}{\textbf{FeSpine3D}} 
& IntraTomo & 0.17  & \underline{29}  & 23.60 & 0.8197 \\
& NeRF      & 0.02  & 466 & \underline{23.86} & \underline{0.8286} \\
& TensoRF   & 0.04  & 152 & 20.79 & 0.6838 \\
& NAF       & \underline{0.23}  & 89 & 23.78 & 0.8271 \\
\rowcolor{gray!10}
& Ours      & \textbf{148.50} & \textbf{7} & \textbf{40.35} & \textbf{0.9815} \\
\hline
\end{tabular}
}
\end{table*}
As shown in Table~1, UniSpine-GS achieves the best reconstruction quality and substantially higher efficiency on both CTSpine3D (X-ray) and FeSpine3D (ultrasound). On CTSpine3D, our method attains 46.54 PSNR and 0.9938 SSIM, surpassing the best-performing baseline (NeRF) by 8.93 dB in PSNR and 0.0136 in SSIM. UniSpine-GS also renders at 113.14 fps, achieving over 300$\times$ speedup compared with NAF (0.37 fps), and converges in 15 min, which is 6.2$\times$ faster than NAF (93 min) under the same setting.

On FeSpine3D, UniSpine-GS attains 40.35 PSNR and 0.9815 SSIM, surpassing the strongest baseline (NeRF) by 16.49 dB in PSNR and 0.1529 in SSIM. It runs at 148.50 fps, achieving an approximately 646$\times$ speedup over the fastest baseline (NAF, 0.23 fps), and converges within just 7 min, which is about 4.1$\times$ faster than IntraTomo (29 min) and 66.6$\times$ faster than NeRF (466 min). Overall, these results consistently demonstrate that UniSpine-GS offers a superior quality–efficiency trade-off for multi-view spine reconstruction across both imaging modalities, enabling high-quality synthesis with fast training and inference.
\subsection{Ablation Study}
\begin{table*}[t]
\centering
\caption{Ablation analysis of our model’s components on CTSpine3D and FeSpine3D datasets. ``$\checkmark$'' and ``$\times$'' denote whether a component is enabled. Best results in \textbf{bold}, while second best \underline{underlined}. We omit the “SPWM-only” setting because SPWM is designed as a loss reweighting module on top of a projection-consistent forward model and stable geometry initialization. Without RIRF and ACUI, the baseline optimization is often unstable and the resulting comparison is less informative.}
\setlength{\tabcolsep}{4.2pt}
\begin{tabular}{llccccc}
\toprule
\textbf{Dataset} & \textbf{Method} & \makecell{\textbf{\hspace{0.8em}RIRF}\\\textbf{+ACUI}} & \textbf{SPWM} & \textbf{PSNR}$\uparrow$ & \textbf{SSIM}$\uparrow$ & \makecell{\textbf{Train}\\\textbf{Time (s)$\downarrow$}}\\
\midrule
\multirow{3}{*}{\textbf{CTSpine3D}} 
& 3DGS        & $\times$      & $\times$      & 37.83 & 0.9856 & 1132 \\
& X-Gaussian  & $\checkmark$  & $\times$      & \underline{46.04} & \underline{0.9926} & \textbf{756} \\
\rowcolor{gray!10}
& UniSpine-GS & $\checkmark$  & $\checkmark$  & \textbf{46.54} & \textbf{0.9938} & \underline{923} \\
\midrule
\multirow{3}{*}{\textbf{FeSpine3D}} 
& 3DGS        & $\times$      & $\times$      & 34.94 & 0.9734 & 850 \\
& X-Gaussian  & $\checkmark$  & $\times$      & \underline{39.91} & \underline{0.9796} & \underline{782} \\
\rowcolor{gray!10}
& UniSpine-GS & $\checkmark$  & $\checkmark$  & \textbf{40.82} & \textbf{0.9846} & \textbf{504} \\
\bottomrule
\end{tabular}
\label{tab:ablation_both}
\end{table*}
As summarized in Table~2, we analyze the contribution of each component on both CTSpine3D (X-ray) and FeSpine3D (ultrasound). On CTSpine3D, introducing the radiative design with RIRF and ACUI markedly improves performance over 3DGS, raising PSNR from 37.83 to 46.04 and SSIM from 0.9856 to 0.9926, while reducing training time from 1132 s to 756 s. This confirms the benefit of a projection-oriented forward model and geometry-aware initialization. Adding the proposed SPWM further improves fidelity, reaching 46.54 PSNR and 0.9938 SSIM. The slightly longer training time of 923 s is mainly due to extra per-iteration computation for constructing the weight map and the finer refinement it encourages near structural boundaries.

Consistent with the above findings, UniSpine-GS also achieves strong gains on FeSpine3D. From the 3DGS baseline, adding RIRF and ACUI boosts PSNR from 34.94 to 39.91 and SSIM from 0.9734 to 0.9796. With the SPWM, PSNR further rises to 40.82 and SSIM to 0.9846, while training time drops from 782 s to 504 s. This suggests the structure-guided weighting is particularly beneficial for FeSpine3D, where projections exhibit lower contrast and more noise-like artifacts, helping stabilize and accelerate optimization.
\subsection{Visualization Analysis}
Qualitative results on both CTSpine3D (X-ray) and FeSpine3D (ultrasound) are shown in Fig.~2, demonstrating that UniSpine-GS produces reconstructions closer to GT with sharper edges, richer mid-/high-frequency textures, and better local detail preservation. In the zoomed regions, competing methods tend to over-smooth vertebral boundaries and wash out subtle intensity variations, while UniSpine-GS maintains clearer structural contours and more consistent textures. Notably, in the ultrasound results of the competing methods, we observe low-contrast whitish artifacts, for example, in the lower region of the NeRF visualization. These artifacts partly explain their lower scores, whereas UniSpine-GS produces cleaner reconstructions with fewer such artifacts.
\begin{figure}[H] 
\centering 
\includegraphics[width=1\textwidth]{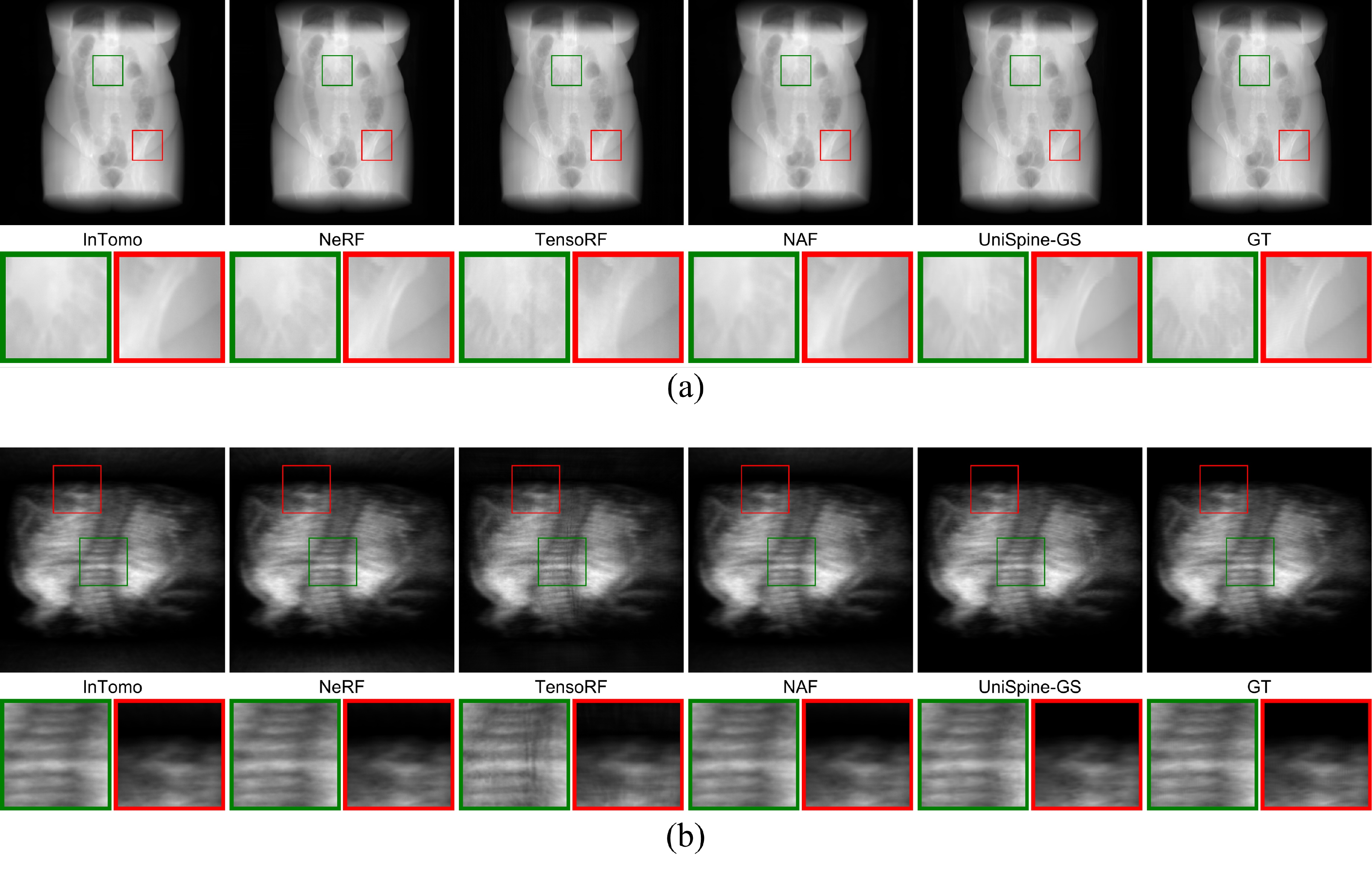}
\caption{Visualization results on (a) CTSpine3D and (b) FeSpine3D datasets.}
\end{figure}
\section{Conclusion}
We propose UniSpine-GS, an efficient physics-aware Gaussian framework for cross-modality multi-view spine image synthesis. With an explicit geometry-aware representation, a unified radiative forward operator, and a parameter-light SPWM, it enhances fine-grained fidelity and cross-view consistency while keeping training and inference fast. Experiments on CTSpine3D and the new FeSpine3D ultrasound dataset show consistent gains over neural field and radiative baselines, achieving a better quality–efficiency trade-off across modalities. Qualitative results demonstrate sharper boundaries and fewer low-contrast artifacts, especially in ultrasound. Future work will explore stronger anatomical priors and lightweight modality adapters, and move toward more ultrasound-consistent differentiable forward models.
\section{Acknowledgments}
This work was supported in part by the National Key Research and Development Program of China under Grant 2023YFC2705702, in part by the National Natural Science Foundation of China under Grants 62225113 and U23B2048, in part by the Innovative Research Group Project of Hubei Province under Grants 2024AFA017, in part by the Science and Technology Major Project of Hubei Province under Grants 2024BAB046 and 2025BCB026, and in part by the New Cornerstone Science Foundation through the XPLORER PRIZE. This work was also supported by WHU-Kingsoft Joint Lab. The numerical calculations in this paper have been done on the supercomputing system in the Supercomputing Center of Wuhan University.

\bibliographystyle{splncs04}
\bibliography{reference}
\end{document}